\documentclass[11pt]{article}
\usepackage{acl-hlt2011}
\usepackage{times}
\usepackage{latexsym}
\usepackage{amsmath}
\usepackage{multirow}
\usepackage{url}
\usepackage{epsfig}
\usepackage{graphicx}

\title{The USTC\_NELSLIP Systems for Trilingual Entity Detection and Linking Tasks at TAC KBP 2016}

\author{Dan Liu$^1$, Wei Lin$^1$, Shiliang Zhang$^2$, Si Wei$^1$, Hui Jiang$^3$ \\
	$^1$iFLYTEK Research, Hefei, Anhui, China\\
	$^2$National Engineering Laboratory for Speech and Language Information Processing \\
	University of Science and Technology of China, Hefei, Anhui, China\\
	$^3$Department of Electrical Engineering and Computer Science \\
	York University,  4700 Keele Street, Toronto, Ontario, M3J 1P3, Canada\\
	{\tt \small \{danliu,weilin2,siwei\}@iflytek.com, zsl2008@mail.ustc.edu.cn, hj@cse.yorku.ca}
}

\date{}

\begin{document}
\maketitle
\begin{abstract}
This paper describes the USTC\_NELSLIP systems submitted to the Trilingual Entity Detection and Linking (EDL) track in  2016 TAC Knowledge Base Population (KBP) contests.
We have built two systems for entity discovery and mention detection (MD): one uses the conditional RNNLM and the other one uses the attention-based encoder-decoder framework. The entity linking (EL) system consists of two modules: a rule based candidate generation and a neural networks probability ranking model. Moreover, some simple string matching rules are used for NIL clustering.
At the end, our best system has achieved an F1 score of 0.624 in the end-to-end typed mention ceaf plus metric.
\end{abstract}

\section{Introduction}
\begin{figure*}[t]
\centering
\epsfig{figure=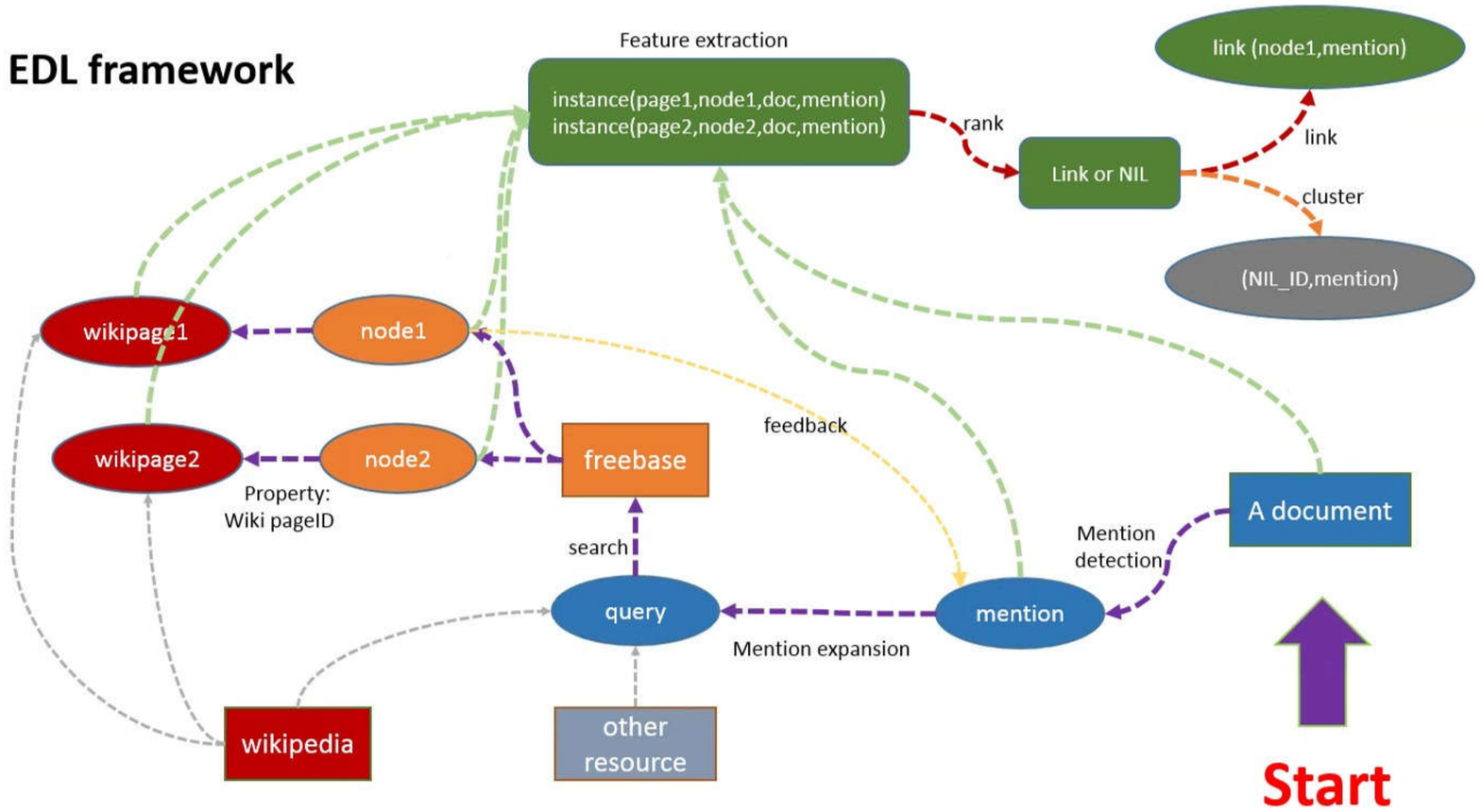,height=8cm}
\caption{{\it The framework of EDL system}}
\label{Fig.edl.framework}
\end{figure*}

In this paper, we describe the USTC\_NELSLIP systems submitted to  2016 TAC KBP Trilingual Entity Discovery and Linking (EDL) task organized by NIST. The EDL task requires to detect named entities and their nominal mentions in the raw text of three languages (English, Chinese and Spanish) and further link each detected mention to the corresponding node in an existing knowledge base, namely Freebase. For NIL mentions that do not exist in the knowledge base, the EDL system needs to cluster all NIL mentions and assign a unique ID to each NIL mention cluster. The entire framework of our EDL systems is shown in Figure \ref{Fig.edl.framework}.

This year, the EDL task has extended the nominal mention detection to all entity types for all three languages. As before, there are in total 5 different mention types, denoted as PER, LOC, ORG, GPE, FAC. During each evaluation window, a large corpus of 90,000 documents is provided to each team to process. Each EDL system needs to be efficient enough to process these documents within the required evaluation window.

\section{Mention Detection}

In the conventional approaches, we normally treat mention detection as a sequence labeling problem, which is typically solved using conditional random fields (CRFs) models. As for nominal mention detection, we typically use a noun phrase chunker to detect all possible candidates for nominal mention. Afterwards, some heuristic post-processing methods are used to identify the true nominal mentions. Differing from the traditional methods, in this work, we consider nominal mentions as special named entities and jointly detect both named and nominal mentions altogether using a single model. In this section, we will describe two different systems built for mention detection: the first system uses RNN-based conditional language model to perform the sequence labeling for mention detection; the other one adopts the popular attention based encoder-decoder structure that is further extended to deal with tree-structured representations to detect nested mentions in the KBP tasks.

\subsection{RNN based Conditional Language Model}

Named entity recognition (NER) without nested entities can be easily formulated as a typical sequence labeling problem, where each output tag can be aligned one by one to an input word.  Moreover, there exists strong dependency among adjacent output labels. Conditional random fields (CRFs) \cite{Lafferty2001conditional} is a widely-used method for sequence labeling. However,  the linear chain CRFs lack of the capability to model long term dependency. We think the long-term dependency may be important to resolve some NER cases. For example, some entity names may get really long, and
the probability for an ORG to occur after a PER (within a certain range) may be quite high.
In the past, some  high order CRFs have been proposed to address these issues but high order CRFs are too complex to train. In this paper,  we introduce a new method to model the long term dependency for NER and mention detection. Let us denote a pair of an input sentence $X$  and an output sequence of tags $Y$ as follows:
%\begin{equation}
\begin{align}
X=(x_1,x_2, ...,x_N) \nonumber \\
Y=(y_1,y_2, ...,y_N). \nonumber
\end{align}
%\end{equation}

Like all  sequence labelling problems, the key problem in modeling is to compute the sequence-level conditional probability $\Pr\left(Y|X\right)$. In this work, we propose a new model to compute the conditional probability as follows:
\begin{equation}\label{eq.2}
\Pr\left(Y|X\right)=\prod_{i=1}^{N} P\left(y_i \; | \; X,y_{i-1},y_{i-2},...y_1\right)
\end{equation}
As shown in eq.(\ref{eq.2}), this modeling approach is quite similar to language models based on recurrent neural networks (RNN)   in \cite{Mikolov2010Recurrent} except that each factorized probability depends on the entire input sequence $X$. Here,  we call this model as conditional RNN language model. The architecture of this models is as shown in Figure \ref{Fig.lable}. In order to compute each factorized conditional probability in  eq.(\ref{eq.2}), we propose to use a hybrid neural network, consisting of two modules. The first model is a convolutional neural network that is stacked with several 1-dimension convolutional layers to generate the representation for the entire input sequence $X$. The second model is a standard RNN-like language model for the output sequence, which always takes the representation of $X$ as input.  For simplicity, we use one layer of gated recurrent units (GRU) \cite{Cho2014On}, which essentially computes all factorized probabilities in eq.(\ref{eq.2}) one by one sequentially, each of which conditions on the CNN-generated representation of $X$ and the preceding partial output sequence.

In the training stage, we jointly learn the CNN layers and the GRU layer to maximize the conditional probability in  eq.(\ref{eq.2}) based on all collected sequence pairs in the training set, $\{ X_i, Y_i\}$. In the test stage, the learned hybrid model of CNNs and GRU-based RNN is used to calculated all conditional probabilities, and the Viterbi decoding algorithm is used to generate the output sequence $Y$ for each input sentence $X$.

%conditioned on the representation of $X$. For the RNN language model, we used one GRU layer
\begin{figure}[h]
\centering
\epsfig{figure=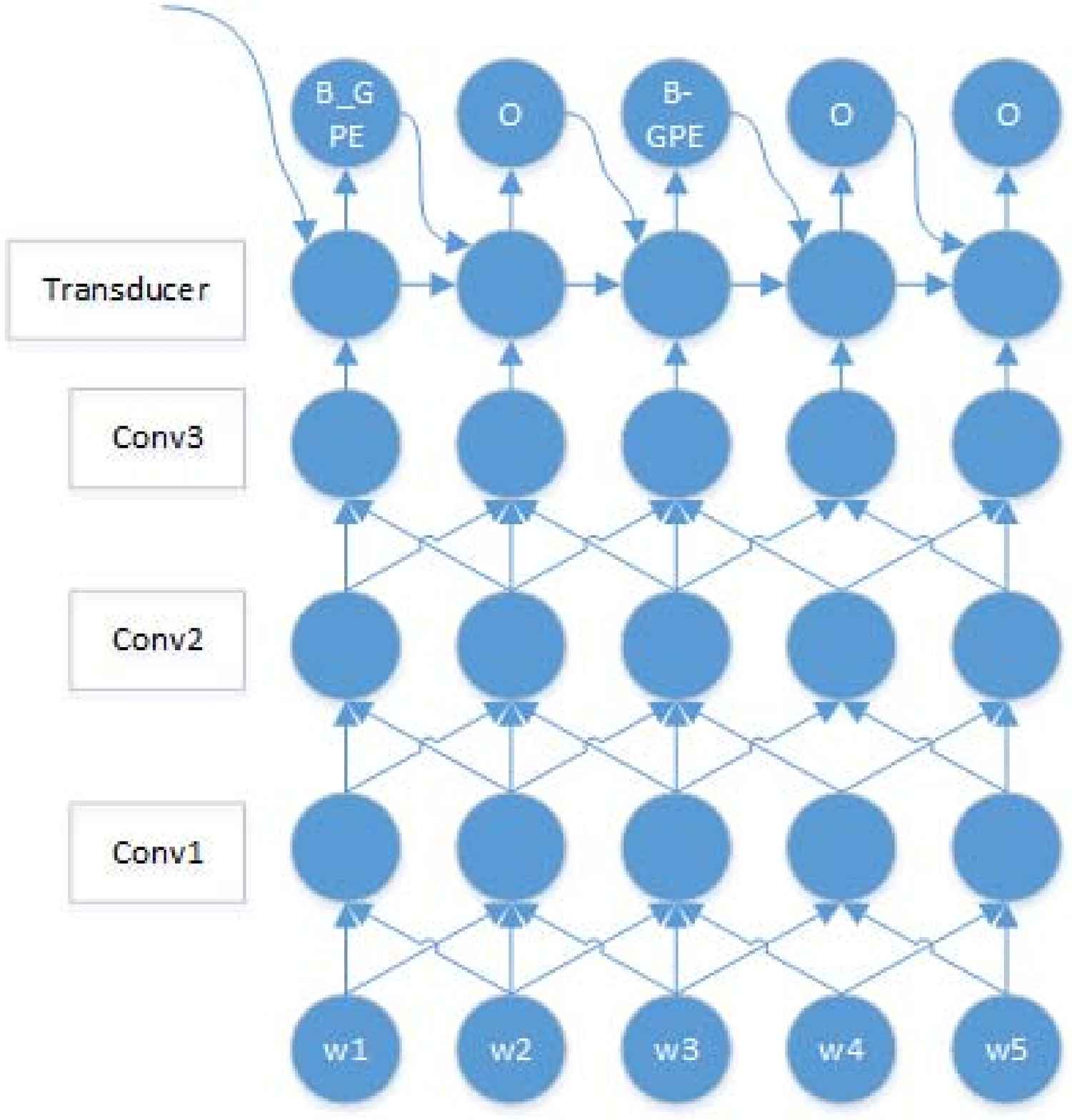,height=7cm}
\caption{{\it The architecture of conditional RNNLM model.}}
\label{Fig.lable}
\end{figure}

\subsection{Attention-based Encoder-Decoder}

In KBP tasks, roughly 10\% of the labelled entities are nested. It is well known it is not straightforward to handle nested entities using the traditional sequence labeling models.
In \cite{finkel2009nested}, it has shown that the nested entities can be processed into a tree-structured representation using a normal char parser. Furthermore, inspired by the idea in \cite{Vinyals2014Grammar}, we may easily linearize a tree structure into a linear sequence representation. For example, given a nested entity such as {\it Kentucky Fried Chicken},  the whole phrase is a named entity of FAC and {\it Kentucky} is a nested named entity of PER. Following the same idea in \cite{Vinyals2014Grammar}, the tree-structured representation for this nested entity may be represented as a linear sequence:
$$[_{FAC} \; [_{PER} \;\;  \mbox{\it Kentucky}  \;  ]_{PER}  \; \mbox{\it Fried} \; \mbox{\it Chicken}  \; ]_{FAC}
$$
where some paired special symbols, such as $[_{FAC}$, $]_{PER}$, $[_{PER}$ and $]_{FAC}$, are introduced to represent the boundaries and type of each entity in a string. Obviously, relying on these extra symbols, this representation is flexible enough to represent any nested entities.

We first use a chart parser to process all entity labels in the training data and generate the corresponding output labels in this format. For example, given an input string like {\it Kentucky Fried Chicken} and its nested entity labels, we will generate the corresponding output sequence as:
$$
[_{FAC} \;\; [_{PER} \;\;  \mbox{\bf Z}  \;\;  ]_{PER}  \;\; \mbox{\bf Z} \;\; \mbox{\bf Z}  \;\; ]_{FAC}
$$
where each $\mbox{\bf Z}$ is a generic placeholder and they correspond to the words of the original sequence one by one in order. Obviously, from this output sequence of placeholder and the input word sequence, we can easily derive all nested entities and their types.

Next, we use an attention-based encoder-decoder model to learn the mapping from the raw word sequence to the above sequence of special symbols and placeholders. The idea is similar to the traditional sequence labelling models, except that the output tages are extended from regular BIO tags to the above special symbols.
The architecture of the attention based encoder-decoder model is shown in Figure \ref{Fig.lable}, which consists of three modules. The encoder module is a stack of several 1-dimension convolutional layers for generating the representation of input sequence $X$, and the attention mechanism is similar to \cite{Bahdanau2014Neural}, and the third module is an RNN-based  decoder to compute the following conditional probability:
\begin{equation}\label{eq.3}
\Pr \left(y_t \; | \; y_1,...,y_{t-1}, X\right)=g\left(y_{t-1}, s_t, c_t \right)
\end{equation}
where $s_t$ is an RNN hidden state at  time instant $t$, $c_t$ is the representation of input $X$ at time instant $t$, and $g\left(\right)$ is a MLP to output conditional probabilities given $y_{t-1}, s_t, c_t$. In this model, we use an attention mechanism to compute $c_t$ as a weighted sum of all input representations $h_t$, where $h_t$ is computed by CNN from the input sentence at time instant $t$.
This attention mmodel works as follows:
\begin{equation}\label{eq.4}
c_t = \sum_{i=1}^{T_x} \alpha_{ti} \; h_i
\end{equation}
where all attention weights $\alpha_{ti}$ is computed by
\begin{eqnarray*}
\alpha_{ti}=\frac{\exp\left(e_{ti}\right)}{\sum_{k=1}^{T_x} \exp\left(e_{tk}\right)} \\
e_{ti}= f\left(s_{t-1}, h_i\right)
\end{eqnarray*}
where $f\left(\right)$ is a MLP to predict attention weights based on $s_{t-1}$ and $h_i$.

The entire model in Figure \ref{Fig.lable} is jointly learned from all training data to maximize the conditional probability of the corresponding output sequence given each input word sentence. In the test stage, for each input word sentence, the learned model is used to compute all conditional probabilities and the Viterbi algorithm is used to generate the output sequence. Occasionally, we may get some unmatched brackets in the output sequences. In these case, we simply drop the unmatched symbols and derive the nested entities based on the remaining part.

\begin{figure}[h]
\centering
\epsfig{figure=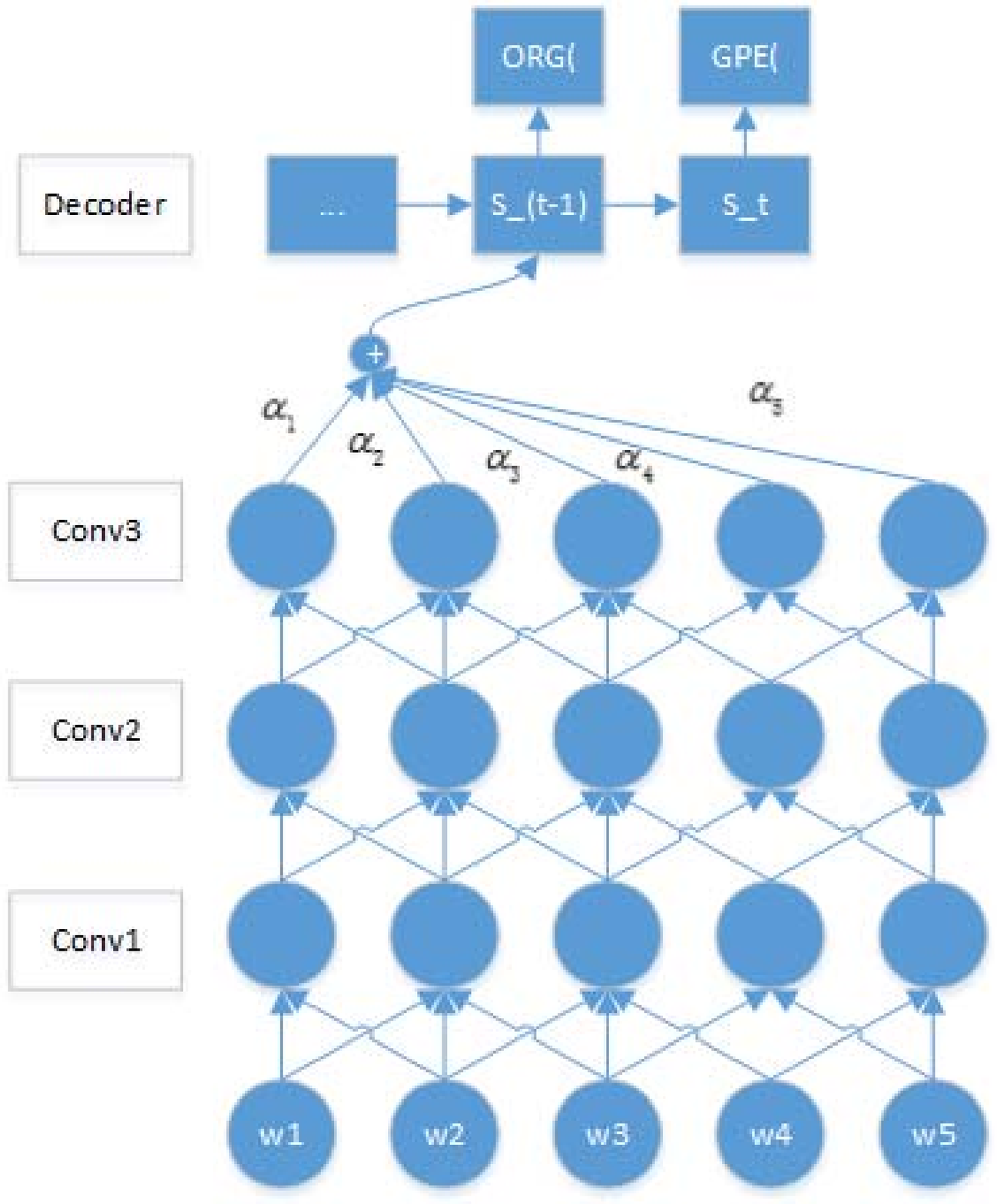,width=10cm}
\caption{{\it The architecture of attention based encoder-decoder.}}
\label{Fig.lable}
\end{figure}

\subsection{Model Configurations}

Both conditional RNN-LM and attention-based encoder-decoder use a stack of five 1-dimension convolutional layers as tne encoder to generate the representations for the input word sequences. In all convolutional layers, we set the filter size and the feature maps to 3 and 512 respectively. We do not use any pooling layers but zero-padding is used in each layer. In this way, the length in each convolutional layer does not change and remain the same as the input sequence. From it, we may easily retrieve the CNN output $h_t$ at every time instant.
In place of the 5-layer CNNs,  we have also examined to use bidirectional GRUs or LSTMs as encoder, but no gain is observed in our experiments. The computation of 1-dimension convolutional layers is much faster than that of RNNs or LSTMs  because of the parallel computation of GPUs. Parameter optimization of all models are performed using AdaDelta \cite{Zeiler2012ADADELTA} and early stopping is also used by monitoring a small held-out development set.

Similar to all neural networks, the performance of our proposed models relies on the amount of the training data. However, there is not too much  matched in-domain training data for the new 2016 KBP mention detection tasks. Therefore, for English and Chinese languages, we have used some in-house data annotated by iFLYTEK research, which consists of about 10,000 Chinese and English documents downloaded from the web. These documents are internally labelled using some annotation rules similar to the KBP guidelines.
For Spanish, we have not found any extra annotated data. Thus, we have trained our Spanish models only using the data from KBP 2015. Because nominal tags are newly introduced to Spanish in KBP 2016, our Spanish models can not predict any nominal tags.

Moreover, we have tried to use model combination to further improve the performance of entity discovery.
For each language, we evenly split all available training data into five parts. For either conditional RNNLM or attention-based encoder-decoder, we have trained 5 different models using only 4 parts of the training data. These models are all randomly initialized. At the end, we use the ensemble of these five models to generate the final entity labels by combining labels scores from these five models.

Finally,  when we use the Viterbi algorithm to generate the output sequence, we implement a  beam search for both models, where we only keep at most 10 active paths at any time instant during the Viterbi decoding.

\section{Entity Linking} \label{section:el}

In the entity linking task, each detected mention needs to be linked to a known entity in an existing knowledge base, namely Freebase in this task. For all mentions that do not match any existing node in Freebase, we need to cluster these NIL mentions. In this work, we adopt a ranking-based  method for entity linking. For a given mention, we first use a rather complicated rule-based system to generate all possible Freebase nodes as the linking targets, each of which is called a linking candidate. This stage is called candidate generation. Next, we train a neural network (NN) based ranking model to rank all these candidates to identity the final linking target. In this step, we have proposed to use many hand-crafted features for the NN-based ranking model.

\begin{figure*}[t]
\centering
\epsfig{figure=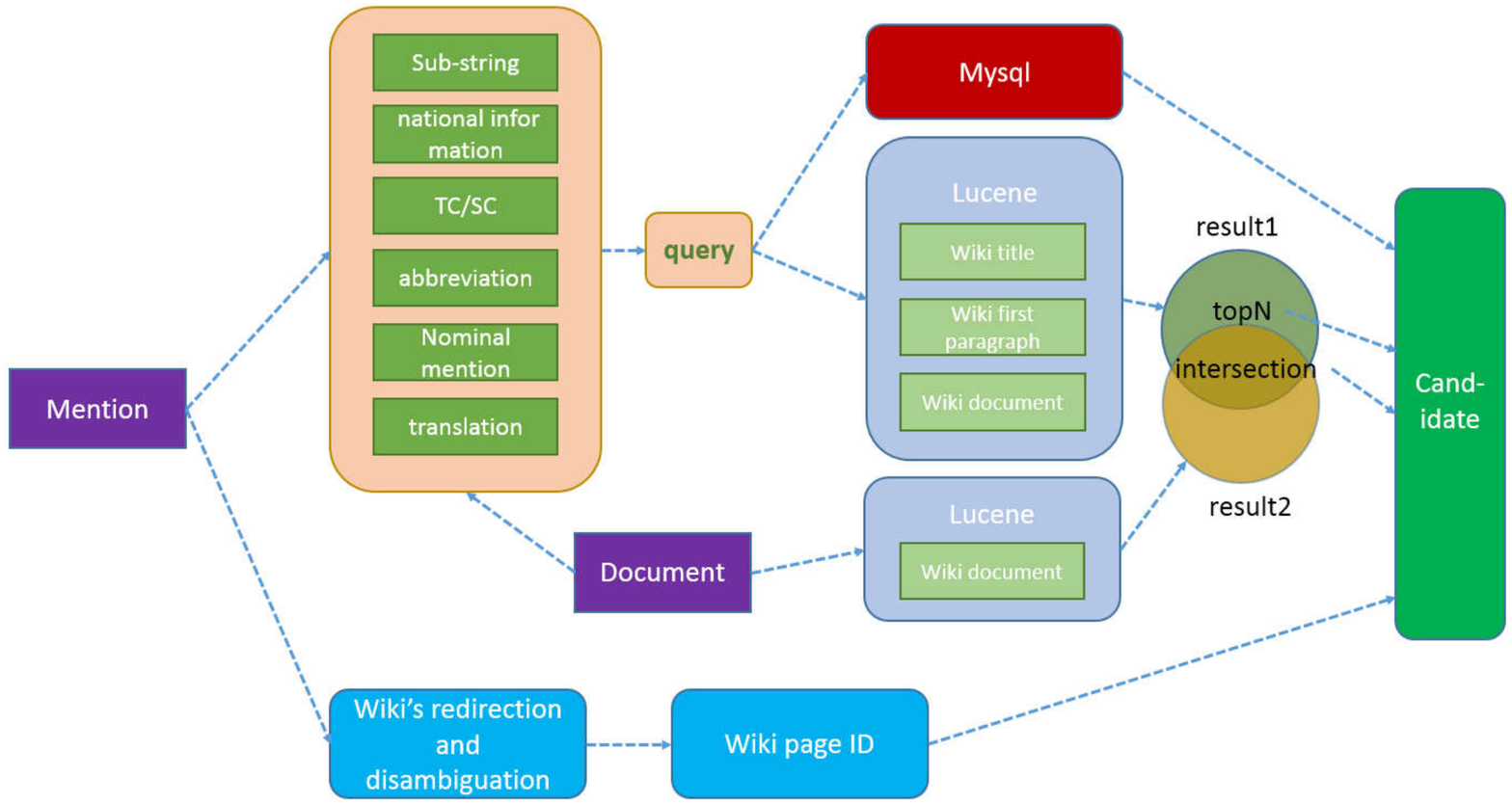,height=8cm}
\caption{{\it The diagram of the entire candidate generation system}}
\label{Fig.candidate}
\end{figure*}

\subsection{Candidate Generation}\label{section:cg}

Obviously, the final linking performance heavily relis on the generated candidate list.
In this work, we have designed a complicated rule-based system as our candidate generation module to generate candidates for each detected mention. The diagram of the whole candidate generation system is shown in Figure \ref{Fig.candidate}.
In this module, candidates are generated based on some knowledge bases, including {\em Freebase}, {\em Wikipedia}.
We have chosen to use Lucene and MySQL for search in our implementation.
The input to this module is a detected mention, the output from this module is a candidate list, which consists of a list of Freebase nodes possibly matching this mention.

In the first step, called {\bf query expansion}, each mention is first expanded into a number of different queries based on some pre-defined rules. These queries represents different ways to rename the same entities. For example, given a detected mention {\em England}, we need to expand it to generate a list of different queries, which may include {\it England}, {\it united kingdom}, {\it united kingdom of great Britain and norther Ireland} and {\it Britain}.
It is better to generate more queries in this step since it may help to enhance the candidate coverage, but this may significantly slow down the following search and matching steps. We need to have a good compromise here.
In this work, we have pre-defined the following rules for the query expansion step:

\begin{enumerate}
	
\item The underlying mention is added to the query list.

\item For each mention, we search the original document containing this mention. If we find this mention is a sub-string of other longer mentions. All of these longer mentions are added to the query list. For instance, if we have a mention like {\it Bush}, and we have found another mention, such as {\it George Bush}, from the same document and {\it Bush} $\subset$ {\it George Bush}, then {\it George Bush} is added to the query list of {\it Bush}.

\item If a mention is in the form of simplified Chinese, its traditional Chinese version is added to the query list, and vice versa.

\item If a mention matches any abbreviations in a pre-compiled list, the corresponding full name is added to the query list. For example, if we have a mention like {\it sc}, we will add {\it South Carolina} to the query list.

\item If a detected mention is nominal, the nearest named mention is selected to go through the above rules 2, 4 to generate the query list. For example, if a nominal mention is detected as {\it president}, its nearest named entity {\em Barack Hussein Obama} is selected for query expansion.

\item If a mention is Chinese or Spanish, we invoke a Google translation API to obtain its English translation. The English translation is used to go through the above rules 2, 4, 5 to expand the query list.

\end{enumerate}

After the query list is ready, we search Freebase nodes and Wikipedia pages to find all possible matches. Since most Wikipedia pages have the corresponding Freebase nodes, we may use these Wikipedia pages as extended context descriptions for the Freebase nodes. We use Lucene and MySQL database to implement search in this step.
MySQL is used to store Freebase and Wikipedia to conduct query searches based on the exact case-insensitive matching.
To improve the recall of the search results, we also need to do fuzzy search and partial matching. For example, if a query is {\it George Bush}, we use the fuzzy search option in Lucene to retrieve the Freebase node labelled as {\it George W. Bush} as well.
To do this, Lucene indexes are built on Freebase nodes titles and Wikipedia pages, fuzzy search is performed using Lucene.
Furthermore, we may directly use the original document containing the underlying mention to search Lucene indexes to generate more results to further improve the coverage of the candidate list.
In our implementation, we first use the expanded queries as input to search Lucene and MySQL to generate the first set of matching results, denoted as {\it Result1}. Next, we use the document as input to search Lucene to generate another set of matching results, denoted as {\it Result2}. Finally, we add the top N  records \footnote{In our experiments, the value of N varies for different languages,
we set N = 3 for English and Spanish, and N = 30
for Chinese.} from {\it Result1} and the intersection of {\it Result1} and {\it Result2} into the list of candidates.
As the final step, for every query, if the query exactly matches (case-insensitive) a title of any Wikipedia page according to the redirection or disambiguation information in Wikipedia, then the Freebase nodes corresponding to these Wikipedia pages are also added to the candidate list.
In order to process those NIL mentions which can not be linked, a special NIL candidate is always added to the candidate list.

\begin{table}[h]
\begin{center}
\begin{tabular}{|l|rrr|}
\hline \bf test set & \bf ENG & \bf CMN & \bf SPA  \\ \hline
coverage &   0.930	 & 0.921 & 0.884 \\
avg. count &   22.60 & 92.96 & 38.55 \\
\hline
\end{tabular}
\end{center}
\caption{\label{Tbl.candidate} Performance of candidate generation on the KBP EDL 2015 dataset for three languages.  }
\end{table}

Here, we use two criteria to measure the qualify of candidate generation: the first one is the total number of different candidates generated for each mention in average (called {\em average count}), and the second one is how many candiate lists actually contain the true target node (called {\em coverage}).
In Table \ref{Tbl.candidate}, we have shown the average count and coverage rate of the candidate lists generated from the above algorithm for three differen languages on the KBP 2015 data set.
In general, our method  generates about 22-100 candidates in average for each mention, varying from one language to another, and the average coverage rates range from 88.4\% (for Spanish) to 93.0\% (for English).

\subsection{Neural Networks Ranking Model}

As described above, we generate a candidate list for each detected mention. This list contains a special NIL candidate and some Freebase node IDs that match with the mention in the candidate generation process. In this work, we have proposed to use a neural network (NN) ranking model to assign probabilities to all candidates in the list. The candidate with the highest probability is chosen as the final linking result. Each time, the NN ranking model takes the mention and a candidate from the list to compute a score. In order to do this, we have designed many handcrafted features for the neural network, which we believe play a decisive role to the final linking performance.

The input feature vector to  the NN ranking model is a concatenation of all the following features:

\begin{enumerate}

\item Mention string embedding (${\bf e}_1$): Each word in the detected mention is projected into a 100-dimension word vector. The sum of all word vectors in the mention is used as the first feature vector, denoted as ${\bf e}_1$.

\item Candidate name embedding (${\bf e}_2$): Each word in the candiate name is also projected into a 100-dimension word vector. The sum of all word vectors in the candidate name is used as another  feature vector, denoted as ${\bf e}_2$.

\item Mention type (${\bf e}_3$): Each mention is represented as a one-hot vector based on the entity type  of the detected mention (PER, ORG, GPE, LOC or FAC). This one-hot vector is projected into a 10-dimension dense vector, denoted as ${\bf e}_3$.

\item Document category (${\bf e}_4$): Each mention is represented as a one-hot vector based on the category of the document containing it (News Report or Discussion Forum). This one-hot vector is projected into another 10-dimension dense vector, denoted as ${\bf e}_4$.

\item Candidate's hot value vector (${\bf e}_5$): A hot value is computed for each candiate based on the number of links the corresponding node has in Freebase. This hot value is quantized into 10 discrete values and represented as a 10-D one-hot vector. This one-hot vector is projected into a 10-dimension dense vector, denoted as ${\bf e}_5$.

\item Edit distance between mention string and candidate name (${\bf e}_6$): A simple edit distance between the mention string and the candiate name is computed as the word numbers. For example, the edit distance between {\it George Bush} and {\it George W. Bush} is 1. The edit distance is quantized and projected into a 10-D vector as above, denoted as ${\bf e}_6$.

\item Cosine similarity of document and candidate description (${\bf e}_7$): Both the document containing the mention and the extended description of the candidate (the corresponding Wikipedia page) are represented as two bag-of-words vectors (normalized by TFIDF). The cosine distance between these two vectors is first computed, and quantized and mapped to a 10-D vector,  denoted as ${\bf e}_7$.

\item Edit distance between translations of mention and candidate (${\bf e}_8$): If the mention or the candidate is Chinese or Spanish, it is translated to English. The edit distance between the English translations of the mention and candidiate is computed, then quantized and projected as ${\bf e}_6$, denoted as ${\bf e}_8$.

\end{enumerate}

%In the EDL2016, we only used the EDL2015 training data set which contains less than 200 documents for training entity linking model for each language. The small training data set is insufficient for statistical model . So handcraft features are the necessary choice. Handcraft features can cover the shortage of data rarity. Model is able to concentrate on linking but feature extraction. Table \ref{Tbl.feature} shows the features designed for the linking model.

\begin{table}[h]
\begin{center}
\begin{tabular}{|c|c|p{5.5cm}|l|}
\hline \bf  & dim & \bf feature  \\ \hline
${\bf e}_1$  & 100 &  mention string embedding \\
${\bf e}_2$  & 100 &   candidate name embedding \\
%2 &   entity type \\
${\bf e}_3$ &  10 & mention type \\
${\bf e}_4$ & 10 &  document type \\
${\bf e}_5$ & 10 &  candidate hot value vector \\
${\bf e}_6$ & 10 &  edit distance between mention string and candidate name\\
${\bf e}_7$ & 10 &  cosine similarity of document and candidate description\\
${\bf e}_8$ & 10 &  edit distance between translations of mention and candidate \\
\hline
\end{tabular}
\end{center}
\caption{\label{Tbl.feature} All input feature vectors used in the NN ranking model. }
\end{table}

\begin{figure*}[t]
\centering
\includegraphics[width=0.5\linewidth]{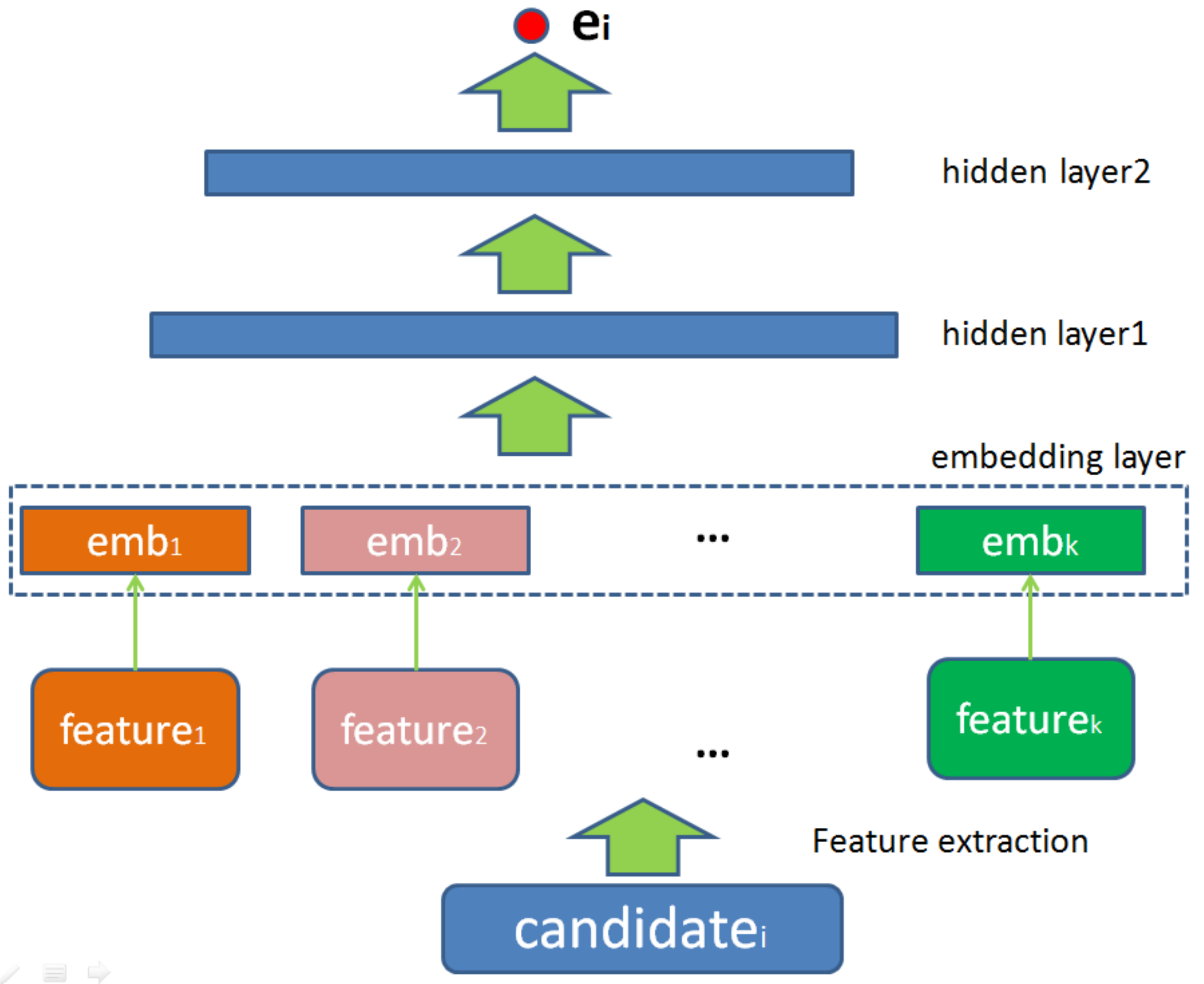}
\caption{The neural network ranking model for entity linking}
\label{Fig:NNRanking}
\end{figure*}

For each detected mention ${\bf m}$, the candidate generation module generates a list of $K$ candidates as $\{ {\bf c}_1, \cdots, {\bf c}_K\}$. For each pair of ${\bf m}$ and ${\bf c}_k$, we generate all feature vectors as shown in Table \ref{Tbl.feature}. These feature vectors are fed into a regular feedforward neural network as shown in Figure \ref{Fig:NNRanking}, to compute a matching score, ${\bf e}_k$. Furthermore, we use a softmax function to compute a posterior distribution of all candidates in the list as follows:

\begin{equation} \label{eq-softmax-ranking}
\Pr({\bf c}_k | {\bf m} ) = \frac{\exp({\bf e}_k)}{\sum_{k=1}^K \exp({\bf e}_k)}.
\end{equation}

In this work, we use the EDL2015 training data set, which contains less than 200 labeled documents to train the neural net ranking model for each language. We choose to use 2 hidden layers: the first layer consists of 512 units while the second layer is composed of 256 units. Each hidden unit use the sigmoid nonlinear activation function. The NN ranking model and all projection matrices in Table \ref{Tbl.feature} are all estimated by maximizing the posterior probabilities in eq.(\ref{eq-softmax-ranking}) of all training data. We adopt a mini-batch AdaDelta with the mini-batch size of 8.
Similar to mention detection, we have also trained 5 different models from different subsets and different random initialization. We have found that an ensemble of five NN ranking yields a small performance gain in the entity linking tasks.

\section{NIL Clustering}\label{section:cluster}

For all mentions identified as NIL by the above NN ranking models, we perform a very simple rule-based
algorithm to cluster them: i) Different named NIL mentions are grouped into one cluster only if their mention strings are the same (case-insensitive); ii) The nominal NIL mention is always grouped to its nearest named mention with the same mention type.
We have investigated other more complex string matching methods for NIL clustering but we have observed no improvement at all.

\section{Experimental Results}

\subsection{Entity Discovery Results}

We have submitted 3 systems to 2016 KBP EDL evaluation. For system 2 and system 3, we use  conditional RNN-LM and attention-based encoder-decoder for entity discovery, respectively. We have observed that these models have achieved a quite high precision but relatively low recall rates. As a result, we have submitted another system by merging the results from the systems 2 and 3. This becomes our top-performing system. The official entity discovery performance from the first EDL1 evaluation in 2016 are summarized  for these three systems in Table \ref{tbl.ed.result}.

\begin{table}[h]
\begin{center}
\begin{tabular}{|l|rrr|}
\hline \bf System & \bf P & \bf R & \bf F  \\ \hline
System2 (cond. RNN-LM) &   0.850	 & 0.678 & 0.754 \\
System3 (attn. enc-dec)           &   0.836  & 0.681 & 0.751 \\
System1 (fusion of 2 and 3)            &   0.822	 & 0.704 & {\bf 0.759} \\
\hline
\end{tabular}
\end{center}
\caption{\label{tbl.ed.result} The official Trilingual Entity Discovery Results of our submitted systems in 2016 KBP EDL1 evaluation. }
\end{table}

\subsection{Entity Linking Experimental Results}

For 2016 KBP EDL evaluation,  we have just developed one entity linking \& NIL clustering system as described in Sections \ref{section:el} and \ref{section:cluster}. Here we just report the official entity linking results from the best entity discovery system (System 1). The performance (in terms of strong\_all\_match) of our system is shown in Table \ref{tbl.el.result} and the performance (in terms of typed\_mention\_ceaf\_plus) is shown in Table \ref{tbl.all.result}.

The results have shown the English system significantly outperform the other two systems. This can be attributed to that the performance of mention detection for English is normally better than the other two language because more English data resources are available in {\it Freebase} and {\it Wikipedia} than Spanish and Chinese.
\begin{table}[h]
%\label{tbl.el.result}
\begin{center}
\begin{tabular}{|l|rrr|}
\hline \bf System & \bf P & \bf R & \bf F  \\ \hline
CMN &   0.692	 & 0.646 & 0.668 \\
ENG &   0.747    & 0.627 & 0.682 \\
SPA &   0.725	 & 0.567 & 0.636 \\
ALL &   0.720	 & 0.617 & {\bf 0.665} \\
\hline
\end{tabular}
\end{center}
\caption{\label{tbl.el.result} The official trilingual entity linking performance of our best system in 2016 KBP EDL evaluation (in terms of strong\_all\_match). }
\end{table}

\begin{table}[h]
%\label{tbl.el.result}
\begin{center}
\begin{tabular}{|l|rrr|}
\hline \bf System & \bf P & \bf R & \bf F  \\ \hline
CMN &   0.658	 & 0.614 & 0.636 \\
ENG &   0.703    & 0.590 & 0.642 \\
SPA &   0.666	 & 0.521 & 0.585 \\
ALL &   0.676	 & 0.579 & {\bf 0.624} \\
\hline
\end{tabular}
\end{center}
\caption{\label{tbl.all.result} The official trilingual entity linking performance of our best system in 2016 KBP EDL evaluation (in terms of typed\_mention\_ceaf\_plus). }
\end{table}

\section{Conclusions}

In this paper, we have described our submitted systems for Trilingual EDL Track of 2016 TAC KBP evaluation.
We have investigated several neural network models for both entity discovery and entity linking. For entity discovery tasks, we have used two neural networks in the popular encoder-decoder framework to model long term dependency and nested entities in the KBP tasks.  For entity linking, we have proposed some handcrafted features and a simple feedforward neural network ranking model. For the NIL clustering, we have adopted a very simple rule-based string-matching clustering method. In overall, our systems have achieved pretty strong performance in both KBP 2015 data and the official KBP 2016 evaluation.

We believe our EDL systems have plenty of room for improvements. For example, we may need to investigate other strategies to detect nominal mentions instead of treating them equally as named entities.
Moreover, some coreference resolution strategies may be used to resolve the relations between entities within the same document or across different documents.

\bibliographystyle{acl}
\bibliography{reference}

\end{document}